\documentclass[conference]{IEEEtran}
\IEEEoverridecommandlockouts

\usepackage{cite}
\usepackage[hidelinks,bookmarks=false]{hyperref}
\usepackage{booktabs} 

\usepackage[cmex10]{amsmath}

\usepackage{amsthm}
\usepackage{algorithmic}
\usepackage{array}
\usepackage{url}

\usepackage[font=footnotesize]{caption}
\usepackage{setspace}
\usepackage{textcomp}
\usepackage{xcolor}

\usepackage{amssymb}
\usepackage{comment}
\usepackage{amsthm}

\usepackage{amsmath}
\usepackage{latexsym}
\usepackage{svg}
\usepackage{float}
\usepackage{stfloats}

\newcommand{\R}{\mathbb{R}}

\usepackage{mathrsfs}
\usepackage[scr=boondoxo]{mathalfa}

\usepackage{lipsum}

\begin{document}

\title{From Volume Rendering to 3D Gaussian Splatting: Theory and Applications}

\newcommand{\cmtid}{233}

\author{
\IEEEauthorblockN{Vitor Pereira Matias\textsuperscript{*}\IEEEauthorrefmark{4}, 
Daniel Perazzo\textsuperscript{*}\IEEEauthorrefmark{2}, 
Vinicius Silva\IEEEauthorrefmark{3}, 
Alberto Raposo\IEEEauthorrefmark{3},}\IEEEauthorblockN{Luiz Velho\IEEEauthorrefmark{2}, Afonso Paiva\IEEEauthorrefmark{4}, and
Tiago Novello\IEEEauthorrefmark{2}}
\IEEEauthorblockA{
\textit{ICMC-USP\IEEEauthorrefmark{4}, IMPA\IEEEauthorrefmark{2}, PUC-RIO\IEEEauthorrefmark{3}}\\
}
}

\maketitle
\begingroup
\renewcommand\thefootnote{*}
\footnotetext{denotes equal contribution.}
\endgroup


\begin{abstract}
The problem of 3D reconstruction from posed images is undergoing a fundamental transformation, driven by continuous advances in 3D Gaussian Splatting (3DGS). By modeling scenes explicitly as collections of 3D Gaussians, 3DGS enables efficient rasterization through volumetric splatting, offering thus a seamless integration with common graphics pipelines. Despite its real-time rendering capabilities for novel view synthesis, 3DGS suffers from a high memory footprint, the tendency to bake lighting effects directly into its representation, and limited support for secondary-ray effects. 
This tutorial provides a concise yet comprehensive overview of the 3DGS pipeline, starting from its splatting formulation and then exploring the main efforts in addressing its limitations. Finally, we survey a range of applications that leverage 3DGS for surface reconstruction, avatar modeling, animation, and content generation—highlighting its efficient rendering and suitability for feed-forward pipelines.
\end{abstract}


\section{Introduction}
3D reconstruction from posed images is a long-standing problem in visual computing that is undergoing a fundamental disruption, driven by advances in Neural Radiance Fields (NeRFs)\cite{mildenhall2021nerf} and 3D Gaussian Splatting (3DGS)\cite{kerbl20233d}.
Given a set of input views with known poses from an unknown scene, the goal is to optimize the parameters of a 3D representation that accurately captures the scene’s geometry and appearance.
NeRFs have had a significant impact on this task by representing scene geometry (as volume density) and radiance using neural networks, optimized through differentiable volume rendering \cite{max1995optical}.
This framework enables highly detailed Novel View Synthesis (NVS) of real-world scenes, with straightforward integration of 3D reconstruction into deep learning pipelines.

\begin{figure}[!h]
    \centering
    \includegraphics[width=\linewidth]{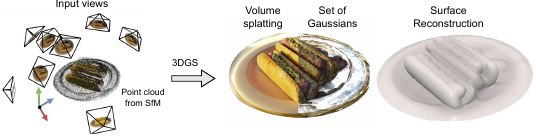}
    \caption{Illustration of the 3DGS pipeline. Given a set of posed input images (left), a sparse point cloud from Structure-from-Motion (SfM) is used to initialize a set of colored 3D Gaussians (middle). These Gaussians are then optimized via volume splatting and can support downstream tasks such as novel view synthesis and surface reconstruction (right).}
    \label{fig:teaser}
\end{figure}

However, representing a scene volumetrically through a neural network can be inefficient, as the network must learn to represent both occupied and empty regions of the domain. During training, its global representation requires supervision across the entire domain, including empty space, resulting in high computational costs and making real-time rendering impractical--crucial for applications.
To address these challenges, Kerbl et al.\cite{kerbl20233d} introduced 3DGS, which models the scene as a collection of 3D Gaussians with colors and leverages volume splatting~\cite{zwicker2002ewa} for differentiable rendering. By avoiding costly queries in empty space and employing rasterization, 3DGS achieves both high detail and real-time performance. Figure~\ref{fig:teaser} gives an overview of 3DGS.

The objective of this tutorial is to present the 3DGS method by deriving its splatting formulation from the volume rendering equation, along with techniques for Gaussian initialization and adaptation during training. We then review recent approaches that address key limitations of the original 3DGS method, including its high memory consumption and limited support for secondary-ray effects. Finally, we discuss a range of applications of 3DGS, including surface reconstruction, animation, avatar modeling, and feed-forward 3D reconstruction from sparse views.
In summary, our contributions are:
\begin{itemize}
\item An intuitive mathematical derivation of 3DGS from the volume rendering equation.
\item A survey of 3DGS extensions and applications across a variety of 3D reconstruction tasks.
\end{itemize}

\section{Volume Rendering Preliminaries}
Our objective is to use differentiable rendering to reconstruct a 3D scene—parameterized by $\theta$—from a set of $N$ posed multi-view images and their corresponding camera intrinsics and extrinsics $\{\mathbf{I}_i, \mathbf{K}_i, \mathbf{W}_i\}_{i=1}^N$.
For each view $i$, we render the scene using the current $\theta$, obtaining an image $\mathbf{I}_{\theta,i}$, which is then compared to its corresponding input view $\mathbf{I}_i$, enforcing $\mathbf{I}_i \approx \mathbf{I}_{\theta,i}$. This is achieved by minimizing the photometric loss:
\begin{align}\label{e-photometric-loss}
\mathcal{L}(\theta) = \frac{1}{N} \sum_{i=1}^N \left\| \mathbf{I}_{\theta,i} - \mathbf{I}_i \right\|^2,  
\end{align}
using standard gradient descent algorithms such as Adam~\cite{kingma2014adam}.

Traditional mesh-based representations are typically non-differentiable, making them unsuitable for optimizing loss \eqref{e-photometric-loss} via gradient descent methods. To overcome this, NeRFs~\cite{mildenhall2021nerf} and 3DGS~\cite{kerbl20233d} employed \emph{volumetric representations} to enable differentiable volume rendering.
Specifically, the scene is parameterized by a density field $\sigma_\theta:\mathbb{R}^3 \to \mathbb{R}$, quantifying particle light absorption, and a color field $c_\theta:\mathbb{R}^3 \to \mathbb{R}$, quantifying emitted radiance. Thus, treating the scene as a ``cloud'' that both absorbs and emits light. Although seemingly unsuitable for representing solid objects initially, this approach has yielded strong results in novel view synthesis\cite{gao2022nerf, chen2024survey} and detailed surface reconstruction~\cite{wang2021neus,huang20242d}.

Let $\gamma(t) = \mathbf{o}_i + t\mathbf{v}$ be the view ray associated with a pixel $\mathbf{p}$ in the input image $\mathbf{I}_i$, where $\mathbf{o}_i$ is the camera position. Our goal is to compute the color observed along this ray using volume rendering techniques~\cite{drebin1988volume}. We restrict the ray domain to the interval $[t_n, t_f]$, where $t_n$ and $t_f$ correspond to the near and far bounds.
Ignoring scattering, the accumulated radiance $I(\gamma(t))$ along the ray satisfies the volumetric light transport ordinary differential equation (ODE)~\cite{max1995optical}:
\begin{align}\label{e-ode}
    I'(t) = \underbrace{\sigma_{\theta}(t)c_{\theta}(t)}_{\text{Emission}} - \underbrace{\sigma_{\theta}(t) I(t)}_{\text{Absorption}}.
\end{align}
For simplicity, we omit $\gamma$ in the notation.
The ODE \eqref{e-ode} can be solved by separation of variables~\cite[Sec. 4]{max1995optical}. Assuming zero background emission as the initial condition, i.e., $I(t_n) = 0$, the accumulated radiance at the end of the ray is given by:
\begin{align}\label{e-volume-rendering}
I_f := I(t_f) = \int_{t_n}^{t_f} \sigma_{\theta}(t) c_{\theta}(t) \exp \left( - \int_{t_n}^{t} \sigma_{\theta}(s) \, ds \right) dt.
\end{align}
The \emph{transmittance} $\exp(\cdots)$ captures the attenuation of light due to absorption along the ray.
In practice, to render the predicted image $\mathbf{I}_{\theta,i}$ for view $i$, we approximate the integral \eqref{e-volume-rendering} using numerical quadrature methods resulting in the volume rendering equation~\cite{mildenhall2021nerf}:
\begin{align}
     \!\!\!I_f \!\approx\!\! \sum_{i=1}^N \!c_\theta(t_i)\! \Big(1 \!-\! \exp\!\big(\!-\sigma_\theta(t_i)\delta_i\big)\Big) \!\prod_{j=1}^{i-1}\!\exp\!\big(\!-\sigma_\theta(t_j)\delta_j\big),
    \label{eq:quadrature}
\end{align}
where $\delta_i = t_i - t_{i-1}$ is the step size between consecutive samples.
This formulation allows gradient-based optimization, as it is fully differentiable. In NeRF, the functions $\sigma_\theta$ and $c_\theta$ are modeled by neural networks, and \eqref{eq:quadrature} is used to optimize $\theta$ via the photometric loss~\eqref{e-photometric-loss}.
However, the global nature of neural representations requires supervision across the entire domain—including empty space—to compute \eqref{eq:quadrature}. This results in high computational overhead and severely limits real-time rendering capabilities, essential for applications.

\section{3D Gaussian Splatting}\label{sec:3dgs}

\begin{figure}[!h]
    \centering
    \includegraphics[width=1\linewidth]{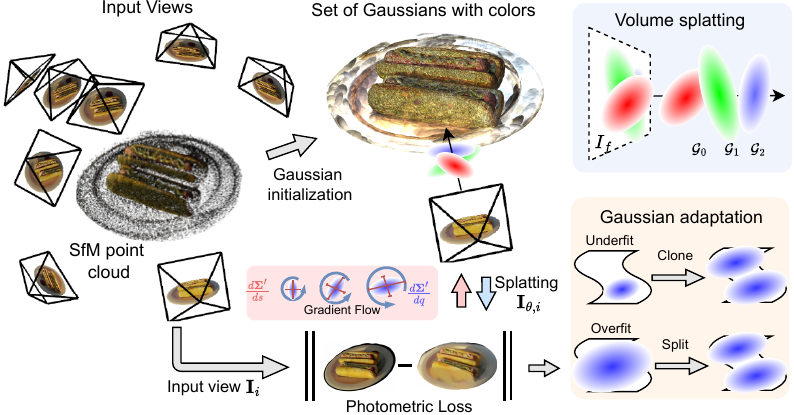}
    \caption{
     Overview of the 3DGS pipeline. The process begins (left) with a set of posed images captured around an object, from which a sparse SfM point cloud is reconstructed. Gaussians are then initialized over this point cloud and optimized (center) through differentiable volumetric splatting. The rendered image is compared to the input views using a photometric loss, whose gradient is used to update the Gaussian parameters. To enhance spatial coverage and avoid under- or over-representation, 3DGS incorporates an adaptation step (right) that dynamically adds (via splitting or cloning) or removes Gaussians during training.
    }
    \label{fig:overview}
\end{figure}

\noindent{\textbf{Gaussian splatting overview}.}
To overcome the computational cost of evaluating the volume rendering equation~\eqref{e-volume-rendering}, Kerbl et al.~\cite{kerbl20233d} introduced 3D Gaussian Splatting (3DGS), which represents the density and radiance fields $\sigma_\theta$ and $c_\theta$ using a collection of colored 3D Gaussians, rendered efficiently via volume splatting~\cite{zwicker2002ewa}.
We follow the pipeline illustrated in Fig.~\ref{fig:overview} to describe the main stages of 3DGS. The input consists of a set of posed images $\{\mathbf{I}_i\}$, from which a colored point cloud is obtained via Structure-from-Motion (SfM)~\cite{schonberger2016structure}; see Fig.~\ref{fig:overview} (top-left). This point cloud serves as the basis for initializing a set of $M$ Gaussians,  
$
g_i := \{\boldsymbol{\mu}_i, \boldsymbol{\Sigma}_i, \sigma_i, c_i\},
$
where \( \boldsymbol{\mu}_i \in \mathbb{R}^3 \) is the Gaussian center, \( \boldsymbol{\Sigma}_i \in \mathbb{R}^{3 \times 3} \) is the covariance matrix, \( \sigma_i \in \mathbb{R} \) is the opacity, and \( c_i \in \mathbb{R}^3 \) is the RGB color. Each Gaussian defines a density function \( \mathcal{G}_i \!:\! \mathbb{R}^3 \!\rightarrow\! \mathbb{R} \) given~by:
\begin{equation}
    \mathcal{G}_i(\mathbf{x}) := \exp \left( -\frac{1}{2} (\mathbf{x} - \boldsymbol{\mu}_i)^T \boldsymbol{\Sigma}_i^{-1} (\mathbf{x} - \boldsymbol{\mu}_i) \right),
\label{e-eval_gauss}
\end{equation}
which controls the spatial influence of the Gaussian around point \( \mathbf{x} \).
To avoid expensive evaluation in empty space, the initial Gaussians can be placed where geometry is present, i.e. the optimization begins with the Gaussian centers $\boldsymbol{\mu}_i$ and colors $c_i$ initialized from the SfM point cloud; see Fig.~\ref{fig:overview}~(top-middle). The scene is then rendered via \emph{volume splatting}, and the output image \( \mathbf{I}_{\theta,i} \) is compared to the corresponding input view \( \mathbf{I}_i \) using a photometric loss. The resulting error is backpropagated to update the Gaussian parameters via gradient descent.
To further refine the representation and avoid poor local minima, an adaptation module is employed during training; see Fig.~\ref{fig:overview} (bottom-right). This step dynamically adjusts the number of Gaussians by splitting Gaussians that are too large, cloning Gaussians that underfit local detail, and pruning those with persistently low opacity.

\vspace{0.1cm}
\noindent{\textbf{Volume splatting}.}
While it is possible to apply quadrature-based volume rendering~\eqref{eq:quadrature} to the Gaussian representation, this is computationally expensive since it would require querying points outside the Gaussian support. Instead, 3DGS adopts \emph{volume splatting}—a efficient rasterization-based alternative—to approximate the integral in \eqref{e-volume-rendering}. 
Precisely, using the view ray $\gamma$ associated to a pixel $\mathbf{p}$, intersecting a set of $K$ Gaussians sorted according to the distance from their centers to the camera position, \autoref{fig:overview}~(top-right).
Using basic properties from integral calculus and assuming that each Gaussian has local support, we can rewrite \eqref{e-volume-rendering} as:
\begin{align}\label{e-general_splat}
\!\!\!I_f \!\approx\!\! \sum_{i=1}^K 
   c_i {\sigma}_i \tau_i \!\prod_{j=1}^{i-1} \!
   \left(1 \!-\! {\sigma}_j \tau_j\right),\text{ with } \tau_i\!=\!\!\int_{\mathbb{R}}\!\mathcal{G}_i\big(\gamma(t)\big)dt.
\end{align}
Zwicker~\cite{zwicker2002ewa} show that $\tau_i$ corresponds to projecting (i.e., \emph{splatting}) the Gaussian into a 2D Gaussian and evaluating it at the pixel location~$\mathbf{p}$. 

To show that \( \int_{\mathbb{R}} \mathcal{G}_i\big(\gamma(t)\big) \, dt \) results in a 2D Gaussian in the image plane, we assume that coordinates are aligned with the camera coordinate system defined by the pose matrix \( \mathbf{W} \). Let \( P_\mathbf{K}: \mathbb{R}^3 \to \mathbb{R}^2 \) denote the perspective projection induced by the camera intrinsics \( \mathbf{K} \). Since \( P_\mathbf{K} \) is non-affine, Zwicker et al.~\cite{zwicker2002ewa} proposed a first-order approximation of \( P_\mathbf{K} \) around the Gaussian center \( \boldsymbol{\mu}_i \) using its Jacobian \( \mathbf{J} \in \mathbb{R}^{2 \times 3} \):
\begin{equation}
    \phi_\mathbf{K}(\mathbf{x}) \approx \phi_\mathbf{K}(\boldsymbol{\mu}_i) + \mathbf{J} (\mathbf{x} - \boldsymbol{\mu}_i).
    \label{eq:linearization}
\end{equation}

This approximation allows computing the integral \( \tau_i \) by evaluating a 2D Gaussian $\tilde{\mathcal{G}}_i: \R^2 \to \R$, called the \textit{splatting} of the 3D Gaussian $\mathcal{G}_i$. To compute the mean and covariance of this Gaussian, we express the 3D Gaussian into the camera coordinate system using $\mathbf{W}$ and apply the linearized projection using first-order approximation:
\begin{equation}
    \tilde{\boldsymbol{\mu}}_i := \phi_\mathbf{K} (\mathbf{W} \boldsymbol{\mu}_i), \quad \quad \mathbf{\tilde{\Sigma}}_i := \mathbf{J W} \mathbf{\Sigma}_i \mathbf{W}^T \mathbf{J}^T.
\end{equation}
The opacity and colors are preserved, that is, $\tilde{\sigma}_i = \sigma_i$ and $\tilde{c}_i = c_i$.
Thus, the resulting \textit{Gaussian splatting} set is given by $\tilde{g}_i = (\tilde{\boldsymbol{\mu}}_i, \tilde{\mathbf{\Sigma}}_i, \tilde{\sigma}_i, \tilde{c}_i,)$. Now, we must perform the composition of multiple volumes with different opacities (alpha-compositing) with the projected Gaussians. 
Assuming the \( K \) Gaussians intersecting the ray ``shooting'' from pixel \( \boldsymbol{p} \in \mathbb{R}^2 \) are sorted by depth, the final light intensity \( I_f \) at that pixel $\boldsymbol{p}$ is given by:
\begin{equation}
\begin{aligned}
I_f 
&\approx \sum_{i=1}^N 
   c_i \,\tilde{\sigma}_i \,\tilde{\mathcal{G}}_i(\boldsymbol{p}) \, \prod_{j=1}^{i-1} 
   \left(1 - \tilde{\sigma}_j \,\tilde{\mathcal{G}}_j(\boldsymbol{p})\right).
\end{aligned}
\label{eq:gaussian_splat}
\end{equation}
With the new intensity equation, the pipeline is reformulated to first project (i.e., splat) the Gaussians onto the image, followed by pixel evaluation. A tile-based \emph{rasterization} strategy enables parallelization for faster volume rendering. Additionally, to reduce aliasing artifacts, 3DGS uses a dilation-based filter~\cite{barron2022mip}. \autoref{fig:overview} provides an overview of the volume splatting operation, where Gaussians are first sorted along the viewing ray and then alpha-composited to form the final image.

\vspace{0.1cm}
\noindent{\textbf{Color representation}.}
In 3DGS~\cite{kerbl20233d}, this modeling approach is used to perform 3D reconstruction from posed images, where the scene parameters $\theta$  are defined as the set of Gaussians.
 They modeled the colors using spherical harmonics to represent view angle light variation and used a diagonalization trick to parameterize the covariance matrix. Since $\mathbf{\Sigma}_i$ is a positive-definite matrix, hence, there is a diagonalization such that $\mathbf{\Sigma}_i:= \mathbf{V} \mathbf{S} \mathbf{V}^T$, where $\mathbf{S} \in \R^{3 \times 3}$ is a diagonal matrix with positive entries and $\mathbf{V} \in SO(3)$, where $SO(3)$ is the 3D rotation group.  
With this representation, we can optimize \( \mathbf{S} \) by directly optimizing its diagonal entries, represented as a vector \( \mathbf{s} \in \mathbb{R}^3 \). For \( \mathbf{V} \), the rotation can be parameterized as an optimizable quaternion \( \mathbf{q} \in \mathbb{R}^4 \).

\section{Extensions and Developments}

The original 3D Gaussian Splatting (3DGS) technique achieves high-quality view synthesis but faces challenges such as memory usage and accurate volume rendering. Recent extensions address these issues through various improvements. Thanks to its strong performance, 3DGS is also being applied beyond its original scope, including surface reconstruction and avatar creation. In this section, we review key developments and highlight notable contributions.

\vspace{0.1cm}
\noindent {\textbf{Memory}:} While possessing high quality, 3DGS employs a substantial quantity of Gaussians, approximately 200,000--500,000 for complex scenes, thereby resulting in significant memory and storage demands, particularly when contrasted with the requirements of NeRFs. Furthermore, these challenges can be exacerbated by the adaptation mechanism, which may introduce additional Gaussians during the training process. Consequently, various methodologies are designed to minimize the quantity of Gaussians, as exemplified by SCAFFOLD~\cite{lu2024scaffold}. This approach employs multi-layer perceptrons (MLPs) to characterize texture attributes and utilizes anchor points to facilitate the distribution of 3D Gaussians. For an exhaustive review of the literature in this domain, refer to 3DGS.zip~\cite{3DGSzip2025}.

\vspace{0.1cm}
\noindent {\textbf{Aliasing and multi-resolution}:} 3DGS uses a dilation filter, which fails to suppress high-frequency artifacts when varying focal length or camera distance. To overcome this, MIP-Splatting~\cite{Yu2023MipSplatting} introduced both 2D and 3D Gaussian filters, improving robustness to aliasing caused by resolution changes. Building on this, recent work has explored Gaussian Splatting for multiresolution applications~\cite{yan2024multi}.

\vspace{0.1cm}
\noindent {\textbf{Specularity}:} In 3DGS, an emission-absorption model~\eqref{e-ode} is used instead of a scattering model~\cite{max1995optical}, effectively baking lighting conditions into the spherical harmonics coefficients. This limits performance on highly reflective surfaces and prevents relighting. To address this, methods such as GaussianShader~\cite{jiang2024Gaussianshader}, 3DGS-DR~\cite{ye20243d}, and IRGS~\cite{gu2025irgs} incorporate classical reflection and shading concepts into the 3DGS framework. An alternative approach by Ginter et al.~\cite{poirier2024diffusion} uses diffusion models for relighting. Additionally, Gao et al.~\cite{gao2024relightable} embed BRDF parameters—albedo, roughness, surface normals, and incident lighting---into each Gaussian, enabling per---point physically based rendering. Direct lighting is reconstructed using environment maps, while spherical harmonics model indirect lighting. Additionally, while standard 3DGS relies only on primary rays, recent methods incorporate secondary rays to model interreflections. To this end, ray tracing techniques have been integrated~\cite{gu2025irgs,3dgrt2024}, allowing secondary rays---such as reflections and refractions---to be spawned upon primary ray interactions with Gaussians.

\vspace{0.1cm}
\noindent {\textbf{Volume Splatting Revisited}:} Volume splatting~\eqref{eq:gaussian_splat} introduces several approximations that can compromise rendering accuracy. Stop-the-pop~\cite{radl2024stopthepop} mitigates popping artifacts caused by the sorting procedure. Other works~\cite{celarek2025does, talegaonkar2025volumetrically} propose compensation terms on the opacity for more accurate volume rendering. Additionally, to improve the affine approximation of the projection operator~\eqref{eq:linearization}, recent methods such as 3DGUT~\cite{wu20253dgut} employ an unscented transform for more precise Gaussian projection onto the image plane.

\vspace{0.1cm}
\noindent {\textbf{3D Reconstruction In the Wild}:} Recent works have extended 3DGS to ``in-the-wild'' settings characterized by occlusions, transient objects, and varying illumination. WildGaussians~\cite{kulhanekwildgaussians} integrates robust DINO features and a lightweight appearance modeling module within the 3DGS framework to handle unconstrained photo collections, matching real-time rendering speeds while outperforming the original NeRF baseline and having similar speed to 3DGS on wild data. 
Gaussian in the Wild (GS-W)~\cite{zhang2024gaussian} separates intrinsic and dynamic appearance attributes per Gaussian point, employs an adaptive sampling strategy, and leverages 2D visibility maps to mitigate transient occlusions and photometric variations, achieving high-fidelity reconstructions at fast render times. In a similar approach, Wild-GS~\cite{xu2024wild} aligns pixel appearance features from image triplanes directly to 3D Gaussians and uses depth regularization and visibility maps to mitigate the transient effects and constrain the geometry. SWAG~\cite{dahmani2024swag} extends 3DGS by conditioning Gaussians on learned appearance embeddings and training unsupervised transient Gaussians to ignore occluders, yielding state-of-the-art results on diverse photo scenes. SpotlessSplats~\cite{sabour2025spotlesssplats} further enhances robustness by detecting outliers in a richer, pre-trained feature space to ignore transient distractors, improving reconstruction quality in casual captures. 

\begin{figure}[ht!]
    \centering
    \includegraphics[width=\linewidth,height=4cm]{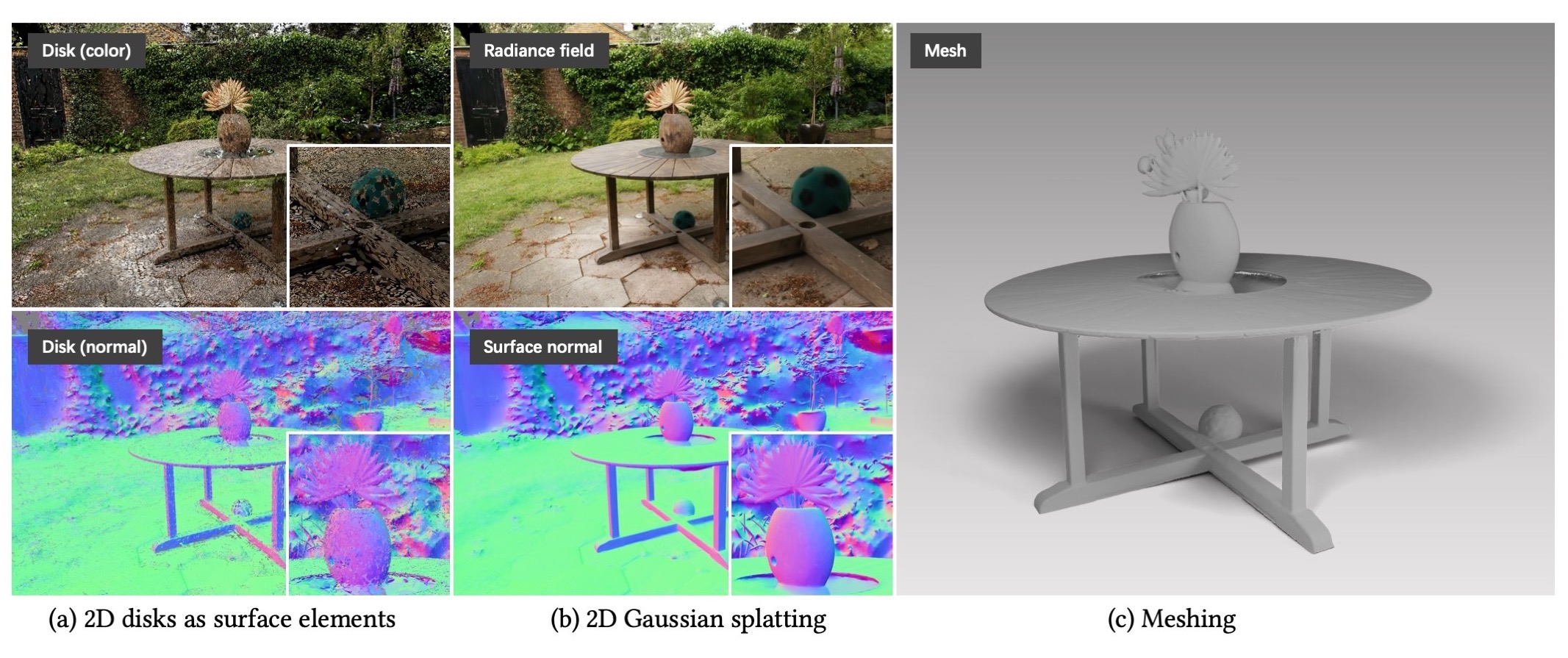}
    \caption{Illustration of 2DGS~\cite{huang20242d}. The method represents the object as a set of 2D disks and successfully recovers both high-quality view synthesis and high-resolution normal maps. Image from Huang et al.~\cite{huang20242d}.}
    \label{fig:surface}
\end{figure}

\vspace{0.1cm}
\noindent{\textbf{Surface reconstruction}:}
Although 3DGS has shown strong performance in NVS, it is not directly suited for mesh extraction. To address this limitation, several techniques have been proposed to augment 3D Gaussian Splatting with surface reconstruction capabilities. One common direction aims to improve the spatial localization of the Gaussians. For instance ``flattening'' the 3D Gaussians~\cite{guedon2024sugar} enables more accurate localization, this was further extended by 2DGS~\cite{huang20242d} which replaced 3D Gaussians with 2D, as shown in \autoref{fig:surface}.
Other works leverage neural signed distance functions (SDFs) for surface reconstruction, inspired by similar approaches used in NeRF-based methods~\cite{wang2021neus, neus2}. Techniques such as GSDF~\cite{yu2024gsdf} and GSPull~\cite{zhang2024gspull} adopt this strategy to recover more detailed mesh structures. Additionally, some recent approaches have proposed rethinking the splatting operation~\cite{yu2024gaussian}.
Several recent approaches have been proposed to address this problem more efficiently, including methods that enable significantly faster reconstruction~\cite{tan2025planarsplatting}, reconstruction from sparse input views~\cite{wu2025sparse2dgs,guedon2025matcha}, and techniques specialized for urban driving scenes~\cite{peng2025desire}.

\vspace{0.1cm}
\noindent {\textbf{Animation}:}
3D representations are often used to enable animation or to recreate dynamic behavior. To achieve physically accurate dynamics, PhysGaussian~\cite{xie2024physgaussian} integrates physics-based continuum mechanics directly within 3DGS, creating a unified simulation-to-rendering pipeline in which each Gaussian is treated as a discrete physical particle with volume, allowing direct simulation. Furthermore, Gaussian Splashing~\cite{feng2025gaussian} performs physical simulations using Position-Based Fluids. To represent fluids more accurately, it builds upon GaussianShader~\cite{jiang2024Gaussianshader}, which enriches Gaussians with material parameters such as diffusiveness, specularity, roughness, and normals to handle specular and reflective objects. To ensure physically accurate simulation, they reconstruct surfaces and disentangle the objects that will be actively simulated, using a mesh-based simulation framework with Gaussians attached to the meshes for texture rendering. Additionally, 4D-GS~\cite{wu20244d} proposes a method for modeling dynamic Gaussians for time-varying content, such as videos. They develop a spatiotemporal structure encoder to capture deformations of both positions and colors over time.

\begin{figure}[ht!]
    \centering
    \includegraphics[width=1\linewidth]{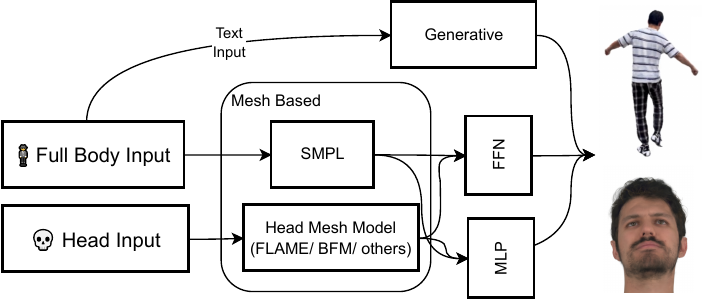}
    \caption{Overview of avatar generation methods. These methods take full-body or head-only inputs, typically processed using body models like SMPL~\cite{smpl} or FLAME~\cite{flame}, respectively. Some approaches use feed-forward networks (FFNs) to generate Gaussians, while others employ MLPs for color effects. Additionally, some methods use text input to guide avatar generation.}
    \label{fig:avatar}
\end{figure}

\vspace{0.1cm}
\noindent{\textbf{Avatars}:}
Reconstructing human avatars using 3DGS poses challenges, such as handling dynamic Gaussians and supporting relighting. To tackle these issues, Gaussian Avatars~\cite{qian2024gaussianavatars} trains a FLAME model~\cite{flame} of the bust, associating each triangle with a Gaussian that is optimized for the scene while being constrained to remain within its corresponding triangle. In contrast, Gaussian Head Avatar (GHA)~\cite{xu2024gaussianheadavatargha} uses two MLPs and a deformation module to predict the 3D Gaussian positions based on facial expression and head pose, alongside a color MLP. Gaussian Avatar~\cite{hu2024gaussianavatar} leverages the SMPL model~\cite{smpl} as a prior to reconstruct full-body avatars using pose features to predict Gaussian parameters through a feed-forward network. Similarly, 3DGS-Avatar~\cite{3dgsavatarqian2024} reconstructs full-body avatars by combining Gaussians with both non-rigid and rigid transformations of an underlying SMPL model. Novel approaches have further advanced avatar generation: GPAvatar~\cite{feng2025gpavatar} achieves high-resolution results, MeGA~\cite{wang2025mega} supports editing capabilities, and DAGSM~\cite{zhuang2025dagsm} enables generative avatars from text descriptions. Additionally, Relightable Gaussian Codec Avatars~\cite{saito2024relightable} introduced relightable Gaussian avatars, supporting more diverse lighting conditions. \autoref{fig:avatar} summarizes these methods for avatar creation. Finally, FLOWING~\cite{bizzi2025flowing} showed a 3D morphing technique for Gaussian Avatars using Flow-based architectures.

\vspace{0.1cm}
\noindent{\textbf{3D reconstruction from sparse views}:}
3DGS~\cite{kerbl20233d} often struggles with NVS from sparse image sets, as its optimization can become trapped in local minima~\cite{charatan2024pixelsplat, zhang2024cor}. To address this challenge, several works have proposed training feed-forward networks (FFNs) to directly predict 3D Gaussian parameters, supervised by photometric losses such as LPIPS, SSIM, or similar metrics, as shown in \autoref{fig:overview}. For instance, Flash3D~\cite{szymanowicz2024flash3d} uses an FFN for monocular reconstruction, generating 3DGS scenes by estimating depth from a single view. PixelSplat~\cite{charatan2024pixelsplat},  incorporates an epipolar transformer to generalize features between two views, pairing this with a probabilistic prediction of Gaussians aligned along camera-to-pixel rays. MVSplat~\cite{chen2024mvsplat} builds upon PixelSplat by combining multi-view feature extraction through transformers and cost volumes, which are then processed by a U-Net. Similarly, GS-LRM~\cite{zhang2024gs} tokenizes two to four images and their camera parameters, concatenates them, and passes them through a transformer block whose output tokens are decoded into Gaussian parameters. NoPoSplat~\cite{ye2024noposplat} addresses critical limitations by eliminating the need for known camera poses. It uses a canonical coordinate system anchored to a single input view and introduces intrinsics token embeddings to resolve scale ambiguity, achieving superior novel view synthesis quality, especially in scenarios with minimal view overlap. Beyond these FFN-based approaches, CoR-GS~\cite{zhang2024cor} improves sparse-view reconstruction by defining two independent reconstructions and comparing them to discard inaccurate Gaussians. Additionally, MAtCha~\cite{guedon2025matcha} employs an atlas-of-charts and depth estimation as input to an MLP, which deforms the atlas and produces high-quality meshes represented with 2D Gaussians.

\begin{figure}[ht!]
    \centering
    \includegraphics[width=\linewidth]{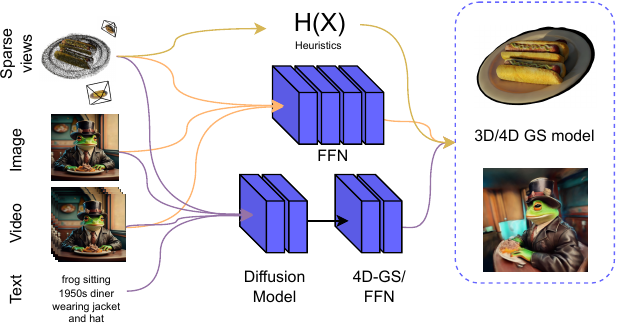}
    \caption{The yellow arrows represent heuristic-based sparse view NVS, e.g. CoR-GS~\cite{zhang2024cor} and MAtCha~\cite{guedon2025matcha}, orange represents FFNs for NVS from sparse views, single image, or video~\cite{charatan2024pixelsplat,chen2024mvsplat,szymanowicz2024flash3d,zhang2024gs,ye2024noposplat}. Finally, purple shows diffusion models being used to generate content from all sorts of inputs~\cite{tang2024lgm,xu2024grm,tang2023dreamgaussian,chen2024lara,wang2024shapemotion4dreconstruction,ren2024l4gm,wu2024cat4dcreate4dmultiview}.
    } 
    \label{fig:feed-forward}
\end{figure}

\vspace{0.1cm}
\noindent{\textbf{Generative models}:} Image or video diffusion models can aid Gaussian Splating as both input and output enhancers. Additionally, 3D content may also be generated with diffusion models that output a set of 3D Gaussians, and can be expanded to 4D animatable objects. LGM \cite{tang2024lgm}, GRM~\cite{xu2024grm}, and DreamGaussian \cite{tang2023dreamgaussian} use multi-view diffusion models as priors 3D generators from text or image inputs, which are passed into a large FFN model that predicts 3D Gaussians to perform 3D reconstruction. The input and architecture of these models vary, as LGM uses an Asymmetric U-Net taking as inputs the MV images with ray embeddings; GRM uses ViT features from the MV images, which are fed into an Upsampler transformer capable of up-sampling the ViT features into a large number of 3D Gaussians. Additionally, LaRa~\cite{chen2024lara} defines a volume transformer taking as input DINO image features from diffusion models or real data to generate 3D content. Extending to 4D content,  Wang et. al. \cite{wang2024shapemotion4dreconstruction}  reconstruct a 4D scene from a single monocular in-the-wild video, using the RGB video plus their respective monocular depths and an additional 2D point tracking over time. L4GM~\cite{ren2024l4gm} also produces 4D scenes from a single monocular video, contrary to Wang et. al., they use diffusion models that take as input a video and generate multi-view videos, which are then passed onto an LGM-like network, resulting in better overall geometry.  CAT4D~\cite{wu2024cat4dcreate4dmultiview} improves results for 4D content generation by extensively using a multi-view video diffusion model from image, video, or text, this diffusion model generates 4D multiview content with a camera-temporal sampling that allows for separate camera and time controls, and allows for time-spatial consistency across all multi-view videos; these are then input into a modified 4D-GS to generate a dynamic 3D scene. \autoref{fig:feed-forward} shows an overview of these methods.

\section{Conclusion and Open Problems}
Gaussian Splatting introduced a novel approach to 3D reconstruction, enabling the creation of high-quality representations. As discussed in this survey, there remain many open problems and promising research directions, such as determining the optimal number of Gaussians, improving the splatting formulation itself, and developing feed-forward 3D reconstruction models that are both fast to use and robust to sparse sets containing any number of input views.

\noindent\footnotesize\textbf{Acknowledgments:} We would like to thank CAPES, as this study was financed in part by the Coordenação de Aperfeiçoamento de Pessoal de Nível Superior – Brasil (CAPES) – Finance Code 001 (grant 88887.842584/2023-00); CNPq; FAPESP; FAPERJ; and Google for partially funding this work.

\bibliographystyle{IEEEtran}
\bibliography{references}

\end{document}